\documentclass{article}
\usepackage{spconf,amsmath,epsfig}

\usepackage{multirow}
\usepackage{multicol}

\let\OLDthebibliography\thebibliography
\renewcommand\thebibliography[1]{
  \OLDthebibliography{#1}
  \setlength{\parskip}{0pt}
  \setlength{\itemsep}{0pt plus 0.3ex}
}

\begin{document}\sloppy

\def\x{{\mathbf x}}
\def\L{{\cal L}}

\title{Region-aware Attention for Image Inpainting}
\name{
Zhilin Huang,
Chujun Qin,
Zhenyu Weng,
Yuesheng Zhu\footnote{Contact Author}
}
\address{
Communication and Information Security Lab, Peking University\\
\{zerinhwang03,chujun.qin,anne\_xin,wzytumbler,zhuys\}@pku.edu.cn
}

\maketitle

\begin{abstract}
Recent attention-based image inpainting methods have made inspiring progress by modeling long-range dependencies within a single image. However, they tend to generate blurry contents since the correlation between each pixel pairs is always misled by ill-predicted features in holes. To handle this problem, we propose a novel region-aware attention (RA) module. By avoiding the directly calculating corralation between each pixel pair in a single samples and considering the correlation between different samples, the misleading of invalid information in holes can be avoided. Meanwhile, a learnable region dictionary (LRD) is introduced to store important information in the entire dataset, which not only simplifies correlation modeling, but also avoids information redundancy. By applying RA in our architecture, our methodscan generate semantically plausible results with realistic details. Extensive experiments on CelebA, Places2 and Paris StreetView datasets validate the superiority of our method compared with existing methods.
\end{abstract}
%

%
\section{Introduction}
\label{sec:intro}

Image inpainting is a task of restoring the missing or damaged parts of images in computer vision. Traditional diffusion-based or patch-based approaches~\cite{DBLP:journals/tog/BarnesSFG09,DBLP:conf/iccv/EfrosL99} only work well for small holes or stationary textural regions, and they fail to generate semantic information on non-stationary images. 

\begin{figure*}[t]
\centering
\includegraphics[width=\textwidth]{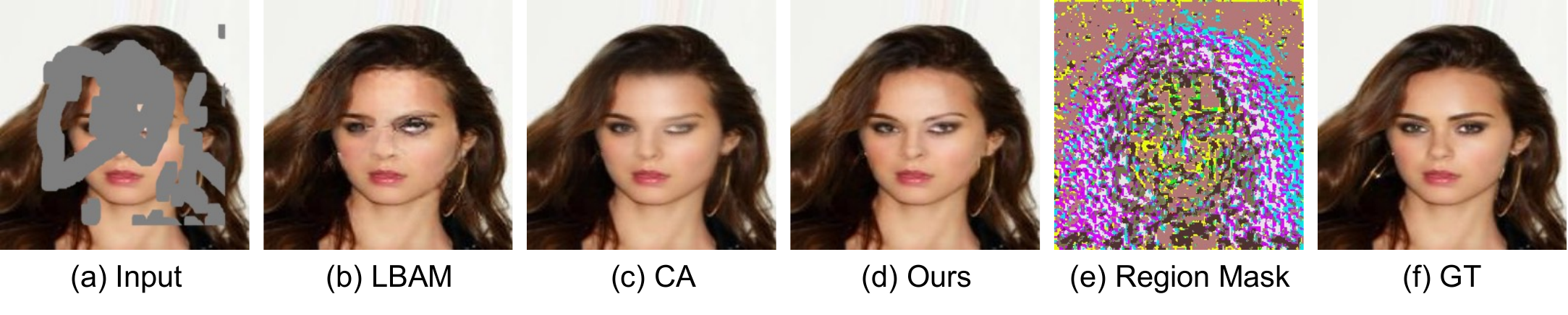}
\caption{Comparison results between LBAM, CA and ours. And (e) presents the region mask generated by the region mask generator in RA. [Best viewed with zoom-in.]}  \label{Fig_Intro}
\end{figure*}

To make up for it, learning-based methods~\cite{DBLP:conf/cvpr/PathakKDDE16,DBLP:conf/iccv/XieLLCZLWD19,DBLP:conf/aaai/YuGJW0LZL20} are proposed to formulate inpainting as a conditional image generation problem by using a vanilla convolutional encoder-decoder network, where the encoder learns a latent feature representation of the image and then the decoder takes the representation to reason about the missing contents. 
Since the receptive fields are limited and the holes in input images are completely empty, convolutions cannot effectively capture the global semantic information of the corrupted images, leading to semantically inconsistent results and artifacts, especially when the holes are large, as shown in Figure~\ref{Fig_Intro}(b) LBAM. 

Recently, attention-based methods are proposed to consider modeling long-range dependencies from a spatial perspective. The contextual attention module (CAM) is proposed in~\cite{DBLP:conf/cvpr/Yu0YSLH18} to enable hole patches to receive information from distant contexts by modeling the relations between contexts and holes. Benefiting from CAM, recent methods~\cite{DBLP:conf/iccv/YuLYSLH19,DBLP:conf/ijcai/WangLZD19} have shown promising performance. 
As CAM models the relation between contextual and holes features patches by computing the cosine similarity, its effectiveness heavily relies on the preliminarily-recovered holes features~\cite{DBLP:journals/corr/abs-Ming-09010}. 
The ill-predicted contents always mislead the modeling of dependencies in CAM, making CAM ineffective in maintaining global semantic consistency in a single image.
Therefore, CAM is often applied in coarse-to-fine architectures which consists of two stacked encoder-decoder networks. In coarse stage, the first encoder-decoder will generate coarse predictions in the damage areas, and then, in fine stage, CAM is applied to the second encoder-decoder to refine the results.
However, using two-stage architectures to obtain valid initial values in holes requires considerable computational resources and makes the networks more difficult to be optimized. 
Besides, CAM only considers to model the correlation within a single sample, ignoring the correlation between different samples.
However, for the inpainting task, it is also necessary to maintain the semantic consistency between samples. 
For example, the unmasked information in sample A can provide clues for the prediction of missing contents in sample B when sample A and B have similar semantics.

In order to effectively model the global dependencies and maintain the semantic consistency in predictions, a novel region-aware attention (RA) module is proposed.
Different from CAM, RA avoids directly calculating the affinity between different positions when modeling the correlation within a single image, and also considers the correlation between different images in the dataset.
Specifically, first of all, a standard convolution is used to map the $c$-dim features of each pixel position in the input feature map to a $n$-dim subspace, where $n$ indicates the number of the regions. 
Then, a share-weight linear layer is applied to the transformed input feature maps for roughly estimating the values in holes of each channel. A guided mask reflecting the coarse correlation between each pixel and each region is obtained.
Since the share-weight linear layer is optimized in an end-to-end manner with the network, it can receive the information from the entire dataset. Therefore, when the share-weight linear layer predicts missing values for each channel, the correlation between different samples can be established which helps the model to avoid the misleading of invalid information in holes. 
Considering the computational burden, we downsample the resolutions of input feature maps by a factor of 4 in practice.
Next, the guided mask is upsampled to the original resolution, and a vanilla convolution followed by softmax is used to refine the relations between each pixel with each region and modeling the relation between each region.
Finally, RA reconstructs missing contents by combining representations of each region which is stored in a learnable region dictionary (LRD), weighted by the correlation between the specific pixel with each region. Since LRD is shared across the entire dataset, it can capture the most important information.
By switching from $N$-to-$N$ ($N$ is the number of the pixels) correlation modeling to $N$-to-$n$ modeling, the process can be simplified and information redundancy can be avoided.
Moreover, For extracting global structure information and local texture effectively, we propose a local-global attention (LGA) layer and apply it to the encoder of our architecture.

Extensive experiments on Places2~\cite{DBLP:journals/pami/ZhouLKO018}, CelebA~\cite{DBLP:conf/iccv/LiuLWT15} and Paris StreetView~\cite{DBLP:journals/cacm/DoerschSGSE15} datasets demonstrate that our method can generate plausible results with clear boundaries and realistic details compared to state-of-the-art methods.

We summarize the contributions of this paper as follow:
\begin{itemize}
\item A novel region-aware attention (RA) module is proposed to indirectly model the correlation between pixels in a single sample from the perspective of region, and consider the semantic consistency between different samples, so as to avoid the misleading of invalid information in the modeling process.

\item In order to maintain the semantic consistency and pay attention to the reconstruction of texture details, we propose a local-global attention (LGA) layer based on RA and apply it to the encoder of our architecture.

\item We show that our approach can achieve promising performance compared with previous state-of-the-art methods on benchmark datasets including CelebA, Paris StreetView and Places2.
\end{itemize}

\section{Related Work}
\subsection{Learning-based Image Inpainting} Learning-based methods for image inpainting~\cite{DBLP:journals/corr/abs-1803-07422,DBLP:conf/iccv/HanWHS019,DBLP:conf/icmcs/LiuGCY019,ma2019coarse,DBLP:conf/nips/WangTQSJ18,xiong2019foreground,DBLP:conf/iccv/XieLLCZLWD19} always use deep learning and adversarial training strategy~\cite{DBLP:conf/nips/GoodfellowPMXWOCB14} to predict the missing contents in hole regions. One of the early learning-based methods, Context Encoder~\cite{DBLP:conf/cvpr/PathakKDDE16} takes adversarial training into a encoder-decoder architecture to fill the holes in feature-level. On the basis of Context Encoder, Iizuka et al.~\cite{DBLP:journals/tog/IizukaS017} propose global and local discriminators to generate better results with regard to overall consistency as well as more detail. Yang et al.~\cite{DBLP:conf/cvpr/YangLLSWL17} propose a multi-scale neural patch synthesis approach to generate high-frequency details. Liu et al.~\cite{DBLP:conf/eccv/LiuRSWTC18} propose an automatic mask generation and update mechanism to focus on valid pixels in the feature map for better results. Inspired by~\cite{DBLP:conf/eccv/LiuRSWTC18}, Yu et al.~\cite{DBLP:conf/iccv/YuLYSLH19} propose a gated convolution and SN-PatchGAN to better deal with irregular masks. Guo et al.~\cite{DBLP:journals/corr/abs-2108-09760} propose a single-stage architecture with two parallel encoder-decoder to process edge information and structure information.

\subsection{Attention-based Image Inpainting} Recently, spatial attention mechanism is introduced in image inpainting task to model long-range dependencies within features~\cite{DBLP:journals/corr/abs-Ming-09010,DBLP:conf/iccv/XieLLCZLWD19}. Yan et al.~\cite{DBLP:conf/eccv/YanLLZS18} introduce a shift operation and a guidance loss to restore features in the decoder by utilizing the information in corresponding encoder layers. Yu et al.~\cite{DBLP:conf/cvpr/Yu0YSLH18} propose a novel contextual attention module to explicitly utilize the feature in known-regions as references to make better predictions. Liu et al.~\cite{DBLP:conf/iccv/LiuJX019} propose a coherent semantic attention layer to model the correlation between adjacency features in hole regions for continuity results. 

\begin{figure*}[t]
\centering
\includegraphics[width=\textwidth]{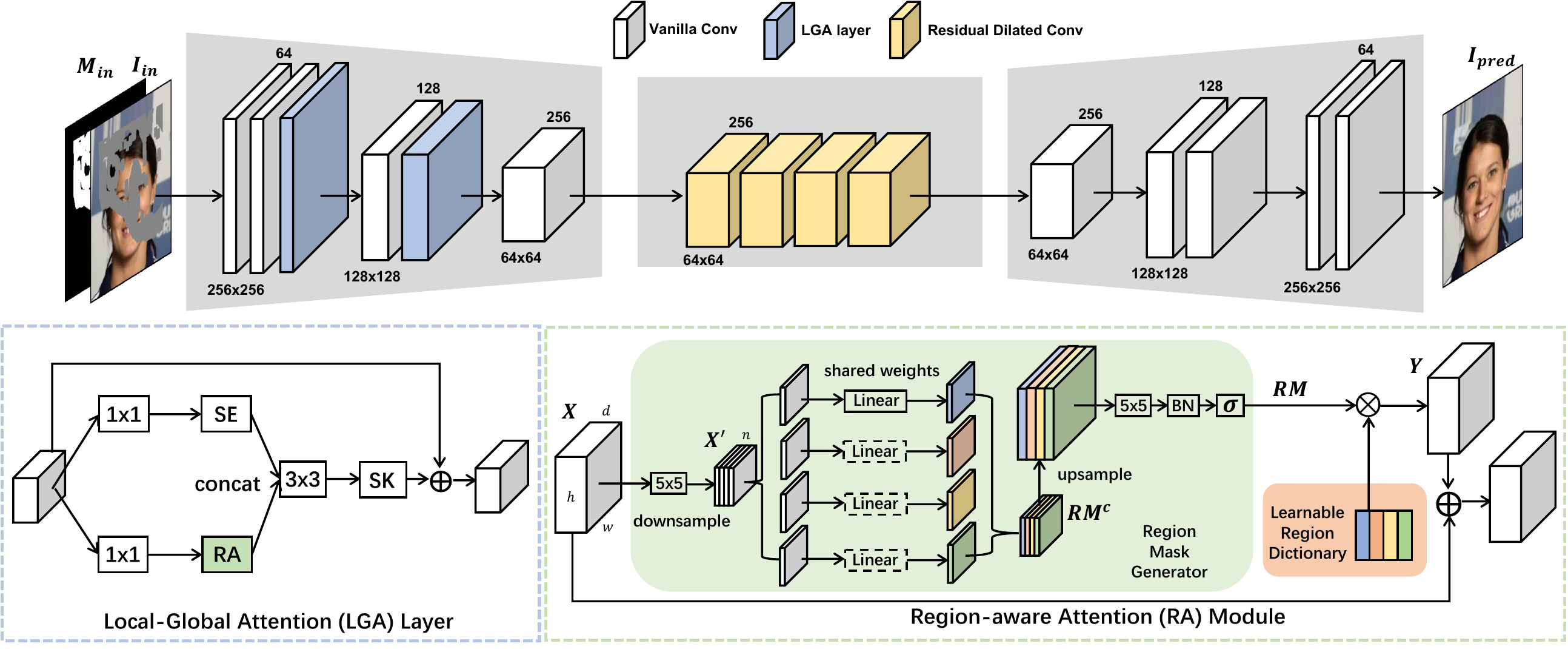}
\caption{Our architecture.} \label{Fig_architecture}
\end{figure*}

\section{Approaches}
\subsection{Region-aware Attention Module}
In order to avoid the misleading of invalid pixels on long-range dependency modeling, we propose a region-aware attention (RA) module.
By introducing a region mask generator (RMG), RA can indirectly model the correlation between pixels in a single image from a region perspective, avoiding the direct influence of invalid pixels.
Meanwhile, the semantic consistency between different samples is maintained by modeling the correlation between samples of entire dataset. 
And a learnable region dictionary (LRD) is introduced to provide important region information for reconstructing the holes under the guidance of RMG.
These gives the model ability to seek clues from other semantically similar samples in the dataset when predicting missing contents in the current sample.

\subsubsection{Region Mask Generator}
For the single-stage architecture, the correlation between each pixel and each region in the input feature map cannot be established directly through the convolution whose receptive fields are limited, since the there are no coarse predictions provided for the empty holes. To handle these, we propose a region mask generator.

For the $c$-dim feature at each pixel position of the input feature map $X \in \textbf{R}^{ c\times h\times w}$, we will use a 5$\times$5 convolution to map them into a $n$-dim subspace, obtaining $X'\in \textbf{R}^{ n\times h\times w}$, where $n$ indicates the number of regions. 
Then, we introduce a share-weight linear layer to channel-wisely predict the coarse values in holes of each channel of $X'$, obtaining the coarse region mask $RM^c \in \textbf{R}^{ n\times h\times w}$.
in practice, $X'$ will be downsampled by 4 times before feeding into the share-weight linear layer for avoiding excessive computational burden, and then the coarse guided mask will be upsampled to the original resolution.
Finally, a convolution is applied to refine the correlation between each pixel with each region reflected in the coarse guided mask and model the correlation between regions, obtaining $RM^r$'. 
And softmax is performed on the channel dimension to obtain the final Region mask $RM$.
The process can be formulated as:
\begin{equation}
    RM^c_i = linear(X'_i) \;
\end{equation}
\begin{equation}
    RM^r = BN(Conv(RM^c)) \; 
\end{equation}\label{func_sigmoid}
\begin{equation}
    RM = softmax(RM^r) \; 
\end{equation}
where $RM^c_i \in \textbf{R}^{(\frac{h}{r} \times \frac{w}{r})} $ and $X'_i \in \textbf{R}^{ (\frac{h}{r}\times \frac{w}{r})} $ indicate $i$-th channel of $RM^c$ and $X'$, respectively. And $r$ is the reduction scale. In practice, we set $r$=4. 

The reason why we utilize a linear layer to generate $RM^c$ instead of a convolution is that convolutions are limited by the receptive field. It is difficult for convolutions to model global information in the spatial dimension, especially in the beginning stage of the network, holes are far larger than the kernel size of convolutions.
At the same time, in order to avoid excessive computation caused by the linear layer, we adopt share-weight linear layer to conduct correlation modeling for the information in each channel.

\subsubsection{Learnable Region Dictionary}
We introduce a learnable $n\times c$ matrix $D$ as a learnable region dictionary (LRD), where the $i$-th row of $D$ stores the important information of $i$-th region. 
And RA reconstructs holes by combining the representations of all regions in LRD according to the region mask generated by RMG.
Specific, for each pixel feature in the input feature maps $X$, RA reconstructs it by combining the representations of all regions in LRD weighted by the elements of $n$-dim vector in the corresponding pixel position of the region mask, which reflects the probability distribution of the current pixel belonging to each region.
\begin{equation}
    Y_i = \sum _{j=1}^n{RM_{ji}\cdot D_j}
\end{equation}
where $Y$ is the reconstructed feature map, $i$ indicates the $i$-th pixel in spatial, $j$ indicates the $j$-th region,  $RM_{ji}$ indicates the element in the $j$-th channel and $i$-th pixel position, $D_{j}$ indicates the $j$-th row of $D$. And the residual connection is introduced to ease the information flow.

Since LRD is shared across the entire dataset and is optimized in an end-to-end manner when the network is optimized by the back-propagation, LRD can learn the most discriminative representations.
Besides, compared with the $N$-to-$N$ modeling manner in CAM, where $N$ is the number of pixels, the $N$-to-$n$ modeling manner in RA only uses the most important information to fulfill the holes, so as to avoid information redundancy.

\subsection{Local-Global Attention Layer}
For helping the model restoring semantically reasonable contents while generating realistic texture details, we propose a local-global attention (LGA) layer.
By introducing two parallel branches, one for using RA to capture global information to ensure the consistency of global semantics, and another one for using SE-Net~\cite{DBLP:conf/cvpr/HuSS18} to capture the relevance of local information. Finally, SK-Net~\cite{Li_2019_CVPR} is introduced to select the necessary information from the fusion of local and global information based on their characteristics. 
The architecture of LGA layer is shown in Figure~\ref{Fig_architecture}.

\subsection{Network Architecture}
We use an encoder-decoder based on vanilla convolutions for single-stage image Inpainting, where both the encoder and decoder are composed of 3 convolutional layers.
And 4 stacked residual dilated convolutional blocks~\cite{gottlob:nonmon,DBLP:journals/corr/YuK15} are embedded between encoder and decoder to enlarge the receptive fields.
The encoder takes the concatenation of the corrupted image $I_{in} \in \textbf{R}^{3\times H \times W}$ and the mask $M_{in} \in \textbf{R}^{H \times W}$ in the channel dimension as input, and the decoder outputs the recovered detailed image $I_{pred} \in \textbf{R}^{3\times H \times W}$.
The LGA layer we proposed is embedded in the first two layers of the encoder.

\subsection{Loss Functions}
To guide the optimization of our model, we introduce image reconstruction loss, perceptual loss, style loss and adversarial loss in our model. 

\subsubsection{Image Reconstruction Loss}
We introduce the image reconstruction loss to supervise the recovery of images.
Specifically, the image reconstruction loss is defined as the $\mathcal{L}_1$ distance between the predicted image and the ground-truth image:
\begin{eqnarray}
\mathcal{L}_{1}=\left\|I_{pred}-I_{gt}\right\|_1 
\end{eqnarray}
where $I_{pred}$ and $I_{gt}$ indicate the predicted image and the ground-truth image respectively.

\subsubsection{Perceptual Loss} For helping the model to capture structural information, we introduce the perceptual loss $\mathcal{L}_{per}$ following~\cite{DBLP:conf/eccv/LiuJSHY20} to the reconstructed detailed image in MS and the structure image in SS. The perceptual loss is defined on the ImageNet-pretrained VGG16:
\begin{eqnarray}
\mathcal{L}_{per}=\mathrm{E}[\sum_{i}\frac{1}{N_i}\left\| \phi_i(I_{pred})-\phi_i(I_{gt})) \right\|_1]
\end{eqnarray}
where $\phi_i$ is the feature map of $i$-th layer of VGG-16. In practice, select the feature map from layers ReLU1\_1, ReLU2\_1, ReLU3\_1, ReLU4\_1, ReLU5\_1 following~\cite{DBLP:conf/eccv/LiuJSHY20}.

\subsubsection{Style Loss} For helping the model to preserve the style coherency, we introduce the style loss $\mathcal{L}_{sty}$ following~\cite{DBLP:conf/eccv/LiuJSHY20}. 
We compute the style loss with given feature map $C_i\times H_i\times W_i$:
\begin{eqnarray}
\mathcal{L}_{sty}=\mathrm{E_i}[\sum_{i}\left\| G^{\phi}_i(I_{pred})-G^{\phi}_i(I_{gt})) \right\|_1]
\end{eqnarray}
where $G^{\phi}_i$ is a $C_i \times C_i$ Gram matrix calculated from the given feature maps.

\subsubsection{Adversarial Loss} The adversarial training strategy is adopted in our work. We follow~\cite{DBLP:conf/iccv/YuLYSLH19} to use SN-PatchGAN, and adopt the Relativistic Average LS adversarial loss~\cite{DBLP:conf/iclr/Jolicoeur-Martineau19} for generating more realistic details:

\begin{small}
\begin{eqnarray}
\mathcal{L}_{adv} = -\mathrm{E}_{I_{gt}} \left[ D(I_{gt},  I_{pred} )^2\right] - \mathrm{E}_{I_{pred}} \left[(1- {D(I_{gt}, I_{pred})}^2)\right] 
\end{eqnarray}
\end{small}


\section{Experiments}
\subsection{Experimental settings}
We evaluate our method on three datasets: CelebA~\cite{DBLP:conf/iccv/LiuLWT15}, Paris StreetView~\cite{DBLP:journals/cacm/DoerschSGSE15} and Places2~\cite{DBLP:journals/pami/ZhouLKO018}. For these datasets, we use the original train, validation and test splits. We train our model on an Nvidia RTX 2080Ti GPU and use Adam algorithm with a learning rate of $1\times 10^{-4}$ for optimization. All masks and images for training and testing are with the size of $256\times256$.
Besides, we adopt the same data augmentation such as flipping during training process as~\cite{DBLP:conf/eccv/LiuJSHY20}.
For fair comparison, we compare our method with 5 existing state-of-the-art methods: LBAM~\cite{DBLP:conf/iccv/XieLLCZLWD19}, RN~\cite{DBLP:conf/aaai/YuGJW0LZL20}, CA~\cite{DBLP:conf/iccvw/NazeriNJQE19}, RT~\cite{DBLP:conf/eccv/LiuJSHY20}, CTSDG~\cite{DBLP:journals/corr/abs-2108-09760}, where LBAM, RN, RT, CTSDG are single-stage methods, and CA is a two-stage method which applies the CAM in the second-stage.
To verify the effectiveness of our model, we make comparisons between ours with these methods in filling irregular holes, which is obtained from~\cite{DBLP:conf/eccv/LiuRSWTC18}.

\subsection{Quantitative Comparisons}
We conduct quantitative comparisons on CelebA, Places2 datasets. We choose mean $L_1$ error, Peak Signal to Noise Ratio (PSNR) and Structural Similarity Index (SSIM) and Fréchet Inception Distance (FID) as evaluation metrics to qualify the performance of these methods. As shown in Table~\ref{Tab_QuantiComp}, our method outperforms all the other methods in filling irregular holes under different hole-to-image area ratios.

\begin{table}[!ht]
\begin{center}
\caption{Quantitative comparison on CelebA, Paris StreetView and Places2 datasets in filling irregular holes with different hole-to-image area ratios. $^-$ Lower is better. $^+$ Higher is better.}
\scalebox{0.8}{
\begin{tabular}{c|c|c|c|c|c|c|c}
\hline
\multicolumn{7}{c}{CelebA}\\
\hline
& Mask& LBAM& RN& CA& RT& CTSDG& Ours\\
\hline
\multirow{4}{*}{$\mathcal{L}_1^-$(\%)} 
& 10-20\% & 0.79& 1.20& 0.77& 0.89& 0.78& \textbf{0.74} \\
& 20-30\% & 1.58& 2.22& 1.50& 1.70& 1.55& \textbf{1.45} \\
& 30-40\% & 2.61& 3.41& 2.45& 2.68& 2.51& \textbf{2.33} \\
& 40-50\% & 3.94& 4.97& 3.61& 3.90& 3.68& \textbf{3.43}\\
\hline
\multirow{4}{*}{$FID^-$} 
& 10-20\% & 6.55& 12.93& 5.73& 7.53& 8.42& \textbf{5.59} \\
& 20-30\% & 13.73& 21.36& 10.93& 13.25& 16.31& \textbf{10.64} \\
& 30-40\% & 21.14& 30.96& 15.60& 18.47& 23.61& \textbf{14.99} \\
& 40-50\% & 31.77& 46.24& 20.23& 22.55& 30.90& \textbf{19.47}\\
\hline
\multirow{4}{*}{$SSIM^+$} 
& 10-20\% & 0.975& 0.964& 0.977& 0.973& 0.974& \textbf{0.979} \\
& 20-30\% & 0.945& 0.932& 0.952& 0.944& 0.946& \textbf{0.955} \\
& 30-40\% & 0.899& 0.887& 0.914& 0.903& 0.904& \textbf{0.920} \\
& 40-50\% & 0.834& 0.819& 0.864& 0.850& 0.849& \textbf{0.872} \\
\hline
\multirow{4}{*}{$PSNR^+$} 
& 10-20\% & 32.29& 30.29& 32.65& 31.59& 32.45& \textbf{33.02} \\
& 20-30\% & 28.33& 27.00& 28.84& 27.93& 28.59& \textbf{29.18} \\
& 30-40\% & 25.42& 24.48& 25.99& 25.40& 25.84& \textbf{26.46} \\
& 40-50\% & 22.95& 22.13& 23.73& 23.26& 23.65& \textbf{24.17} \\
\hline
\hline
\multicolumn{7}{c}{Places2}\\
\hline
& Mask& LBAM& RN& CA& RT& CTSDG& Ours\\
\hline
\multirow{4}{*}{$\mathcal{L}_1^-$(\%)}
& 10-20\% & 1.28& 1.21& 1.07& 1.09& \textbf{1.03} & \textbf{1.03} \\
& 20-30\% & 2.26& 2.19& 1.99& 2.18& \textbf{1.86} & 1.90\\
& 30-40\% & 3.23& 3.30& 2.98& 3.30& 2.78 & \textbf{2.58} \\
& 40-50\% & 4.38& 4.60& 4.17& 4.66& \textbf{3.83} & 4.00\\
\hline
\multirow{4}{*}{$FID^-$} 
& 10-20\% & 18.93& 31.75& 16.51& 20.71& 19.23& \textbf{15.79} \\
& 20-30\% & 30.21& 46.83& 27.73& 33.62& 31.29& \textbf{26.15} \\
& 30-40\% & 41.98& 63.51& 37.72& 44.86& 44.08& \textbf{35.81} \\
& 40-50\% & 52.96& 80.76& 47.62& 55.51& 59.05& \textbf{45.48} \\
\hline
\multirow{4}{*}{$SSIM^+$} 
& 10-20\% & 0.928& 0.916& 0.939& 0.931& 0.939& \textbf{0.943} \\
& 20-30\% & 0.866& 0.862& 0.883& 0.867& 0.887& \textbf{0.889} \\
& 30-40\% & 0.798& 0.807& 0.818& 0.794& \textbf{0.826}& \textbf{0.826} \\
& 40-50\% & 0.713& 0.737& 0.737& 0.701& \textbf{0.751} & 0.747\\
\hline
\multirow{4}{*}{$PSNR^+$} 
& 10-20\% & 28.85& 28.47& 29.89& 28.97& 30.13& \textbf{30.16} \\
& 20-30\% & 25.84& 26.07& 26.65& 25.85& 27.08& \textbf{27.32} \\
& 30-40\% & 23.81& 24.05& 24.34& 23.60& 24.93& \textbf{25.06} \\
& 40-50\% & 22.20& 22.35& 22.50& 20.80& 23.24& \textbf{23.08} \\
\hline
\end{tabular}}
\end{center}
\label{Table_irr}
\end{table}

\begin{figure*}[!h]
\centering
\includegraphics[width=12.5cm]{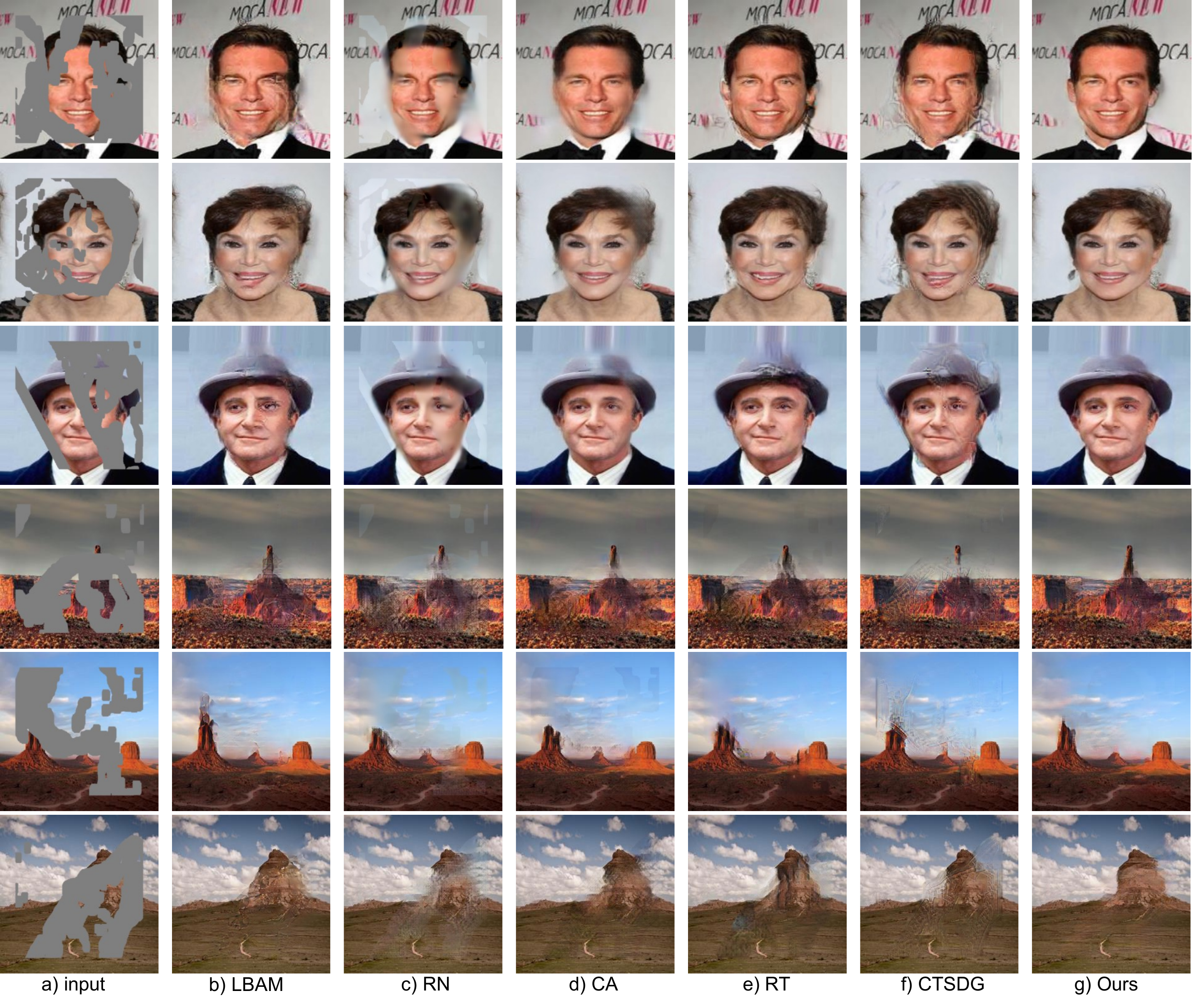}
\caption{Qualitative comparison results on CelebA and Places2 datasets with irregular holes. [Best viewed with zoom-in.]} \label{Fig_QualiComp1}
\end{figure*}

\subsection{Qualitative Comparisons}
We conduct qualitative comparisons on CelebA and Places2 datasets which contain human faces and natural scenes.
Figure~\ref{Fig_QualiComp1} presents the comparison results in filling irregular holes. 
The results of RN and CA present that they tend to generate distorted structure and blurry textures. And LBAM and CTSDG tend to generate artifacts. Although results produced by RT have more reasonable structures, there still exists artifacts and blurry boundaries. Compared with these methods, our model is able to synthesize semantically reasonable results with realistic details.

\subsection{Ablation Study}\label{sec_abla}
\subsubsection{The Effect of LGA Layer}
In order to verify the effectiveness of LGA and demonstrate the advantages of our method compared with CAM in single-stage architectures, we conduct 4 experiments on Paris StreetView dataset:
\begin{itemize}
    \item \textbf{Exp 1}: We remove LGA layers embedded in the encoder of our architecture, obtaining the BasicNet.
    \item \textbf{Exp 2}: We replace the RA in the LGA layers embedded in the encoder of our architecture with CAM.
    \item \textbf{Exp 3}: We embed LGA layer in the last two decoder layers of BasicNet, and replace the RA in the LGA layers with CAM.
    \item \textbf{Exp 4}: Our model.
    \item \textbf{Exp 5}: We embed LGA layer in the last two decoder layers of BasicNet. 
\end{itemize}
By comparing Exp2 and Exp3, we find that applying CAM in decoder of single-stage architecture achieves better performance than applying CAM in Encoder.
Meanwhile, by comparing Exp1 and Exp2, we find that applying CAM in the encoder of single-stage architecture brings limited improvement.
These prove that CAM is sensitive to the validity of the features in holes, especially when the holes are completely empty.
The comparison between Exp1 and Exp4 validate the effectiveness of LGA layer we propose.
The comparisons between Exp4 with Exp2 and 3 present that our RA can effectively model the correlation between each pixel with each region in the single-stage architecture, avoiding the misleading of invalid features in holes.
We attribute this characteristics to the share-weight linear layer in RA, because it provides a coarse prediction for holes from a global perspective.
Through the comparison between Exp4 and Exp5, we find that applying LGA to encoder can bring better performance.
The reason may be that providing appropriate initial values for holes via the LGA at the beginning of the network allows more subsequent components pay attention to refine the details.
\begin{table}[htbp]
\centering
\caption{The effect of LGA layer. $^-$Lower is better. $^+$Higher is better.}
\scalebox{1.0}{
\begin{tabular}{|c|cccc|}
\hline
\textbf{Exp} & \textbf{$\mathcal{L}_1^-$(\%)}& FID$^-$& SSIM$^+$& PSNR$^+$\\
\hline
\textbf{1} & 3.22& 61.82& 0.829& 24.99\\
\hline
\textbf{2} & 3.05& 52.44& 0.839& 25.31\\
\hline
\textbf{3} & 2.96& 49.66& 0.844& 25.59\\
\hline
\textbf{4} & \textbf{2.84}& \textbf{45.48}& \textbf{0.857}& \textbf{26.03}\\
\hline
\textbf{5} & 2.89& 49.74& 0.851& 25.87\\
\hline
\end{tabular}
}

\label{Tab_abla_LGA}
\end{table}

\subsubsection{The Visualization of Region Mask}
\begin{figure*}[h]
\centering
\includegraphics[width=\textwidth]{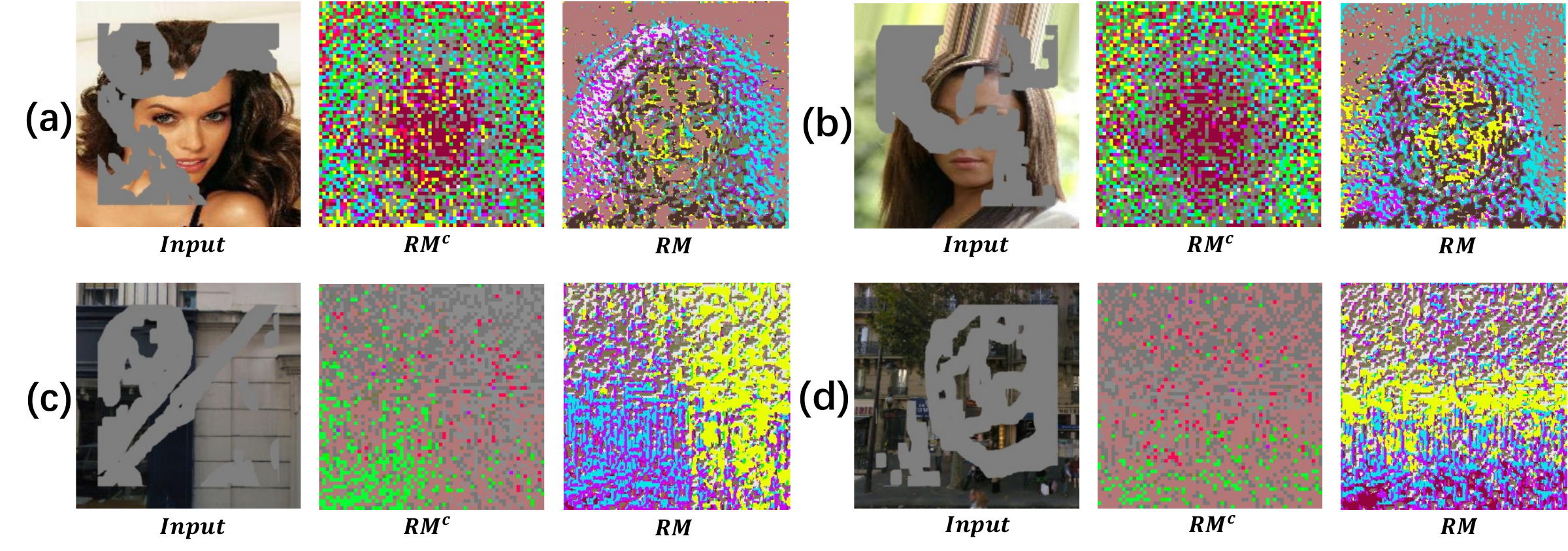}
\caption{The visualization of region mask. [Best viewed with zoom-in.]} \label{Fig_visualization}
\end{figure*}

We visualize the region mask $RM$ generated in RA, as shown in Figure~\ref{Fig_visualization}. 
For the human-face dataset CelebA, We found that the region mask generator (RMG) is able to predict the missing values and divide the regions based on human facial characteristics. 
However, for the natural scenes dataset Paris StreetView, RMG can only roughly divide the image into foreground and background.
We think the reason is that the natural scenes are more complicated and we don't have any supervision about the region division during the training process, so the model can only learn some general regional information.
Meanwhile, we visualize the prediction results $RM^c$ of share-weight linear layer of RMG. The comparisons between $RM^c$ and $RM$ present that the convolution following the linear layer can refine the predicted region mask.

\subsubsection{The Effect of Learnable Region Dictionary}
Compared with the $N$-to-$N$ modeling manner in CAM, where $N$ is the number of pixels, the $N$-to-$n$ modeling manner in RA only uses the most important information stored in learnable region dictionary to fulfill the holes. 
To explore the influence of the different $n$ values in LRD on the performance of the final model, we conduct the experiments on CelebA dataset, as shown in Table~\ref{Tab_abla_n}.
The results presents that when $n$ in LRD is set to 16, the model can achieve the best performance on CelebA dataset.
\begin{table}[htbp]
\centering
\caption{The effect of different $n$ in LRD. $^-$Lower is better. $^+$Higher is better.}
\scalebox{1.0}{
\begin{tabular}{|c|cccc|}
\hline
\textbf{Exp} & \textbf{$\mathcal{L}_1^-$(\%)}& FID$^-$& SSIM$^+$& PSNR$^+$\\
\hline
\textbf{n=8} & 2.44& 15.85& 0.917& 26.36\\
\hline
\textbf{n=16} & \textbf{2.33}& \textbf{14.99}& \textbf{0.920}& \textbf{26.46}\\
\hline
\textbf{n=32} & 2.35& 15.34& 0.918& 26.37\\
\hline
\end{tabular}
}
\label{Tab_abla_n}
\end{table}

\section{Conclusions}
In order to solve the problem that attention is susceptible to invalid information and cannot effectively model long-range dependency within a single corrupted image, we propose a novel region-aware attention for image inpainting.
By avoiding the directly calculating corralation between each pixel pair in a single sample and considering the correlation between different samples, the negative impact of invalid information in modeling global correlation can be avoided.
Meanwhile, a learnable region dictionary (LRD) is introduced to store important information in the entire dataset, which not only simplifies correlation modeling, but also avoids information redundancy.
By applying RA in the encoder of our architecture, our methods can generate semantically plausible results with realistic details. 
Extensive experiments validate the superiority of our method compared with existing methods.

\section{Acknowledgement}
This work was supported in part by the National Innovation 2030 Major S\&T Project of China under Grant 2020AAA0104203, in part by the Nature Science Foundation of China under Grant 62006007, and in part  by the Shenzhen Municipal Development and Reform Commission (Disciplinary Development Program for Data Science and Intelligent Computing).

\bibliographystyle{IEEEbib}
\bibliography{myarxiv}

\begin{thebibliography}{10}

\bibitem{DBLP:journals/tog/BarnesSFG09}
Connelly Barnes, Eli Shechtman, Adam Finkelstein, and Dan~B. Goldman,
\newblock ``Patchmatch: a randomized correspondence algorithm for structural
  image editing,''
\newblock {\em {ACM} Trans. Graph.}, vol. 28, no. 3, pp. 24, 2009.

\bibitem{DBLP:conf/iccv/EfrosL99}
Alexei~A. Efros and Thomas~K. Leung,
\newblock ``Texture synthesis by non-parametric sampling,''
\newblock in {\em {ICCV}}. 1999, pp. 1033--1038, {IEEE} Computer Society.

\bibitem{DBLP:conf/cvpr/PathakKDDE16}
Deepak Pathak, Philipp Kr{\"{a}}henb{\"{u}}hl, Jeff Donahue, Trevor Darrell,
  and Alexei~A. Efros,
\newblock ``Context encoders: Feature learning by inpainting,''
\newblock in {\em {CVPR}}. 2016, pp. 2536--2544, {IEEE} Computer Society.

\bibitem{DBLP:conf/iccv/XieLLCZLWD19}
Chaohao Xie, Shaohui Liu, Chao Li, Ming{-}Ming Cheng, Wangmeng Zuo, Xiao Liu,
  Shilei Wen, and Errui Ding,
\newblock ``Image inpainting with learnable bidirectional attention maps,''
\newblock in {\em {ICCV}}. 2019, pp. 8857--8866, {IEEE}.

\bibitem{DBLP:conf/aaai/YuGJW0LZL20}
Tao Yu, Zongyu Guo, Xin Jin, Shilin Wu, Zhibo Chen, Weiping Li, Zhizheng Zhang,
  and Sen Liu,
\newblock ``Region normalization for image inpainting,''
\newblock in {\em {AAAI}}. 2020, pp. 12733--12740, {AAAI} Press.

\bibitem{DBLP:conf/cvpr/Yu0YSLH18}
Jiahui Yu, Zhe Lin, Jimei Yang, Xiaohui Shen, Xin Lu, and Thomas~S. Huang,
\newblock ``Generative image inpainting with contextual attention,''
\newblock in {\em {CVPR}}. 2018, pp. 5505--5514, {IEEE} Computer Society.

\bibitem{DBLP:conf/iccv/YuLYSLH19}
Jiahui Yu, Zhe Lin, Jimei Yang, Xiaohui Shen, Xin Lu, and Thomas~S. Huang,
\newblock ``Free-form image inpainting with gated convolution,''
\newblock in {\em {ICCV}}. 2019, pp. 4470--4479, {IEEE}.

\bibitem{DBLP:conf/ijcai/WangLZD19}
Ning Wang, Jingyuan Li, Lefei Zhang, and Bo~Du,
\newblock ``{MUSICAL:} multi-scale image contextual attention learning for
  inpainting,''
\newblock in {\em {IJCAI}}. 2019, pp. 3748--3754, ijcai.org.

\bibitem{DBLP:journals/corr/abs-Ming-09010}
Yong{-}Goo Shin, Min{-}Cheol Sagong, Yoon{-}Jae Yeo, Seung{-}Wook Kim, and
  Sung{-}Jea Ko,
\newblock ``{PEPSI++:} fast and lightweight network for image inpainting,''
\newblock {\em CoRR}, vol. abs/1905.09010, 2019.

\bibitem{DBLP:journals/pami/ZhouLKO018}
Bolei Zhou, {\`{A}}gata Lapedriza, Aditya Khosla, Aude Oliva, and Antonio
  Torralba,
\newblock ``Places: {A} 10 million image database for scene recognition,''
\newblock {\em {IEEE} Trans. Pattern Anal. Mach. Intell.}, vol. 40, no. 6, pp.
  1452--1464, 2018.

\bibitem{DBLP:conf/iccv/LiuLWT15}
Ziwei Liu, Ping Luo, Xiaogang Wang, and Xiaoou Tang,
\newblock ``Deep learning face attributes in the wild,''
\newblock in {\em {ICCV}}. 2015, pp. 3730--3738, {IEEE} Computer Society.

\bibitem{DBLP:journals/cacm/DoerschSGSE15}
Carl Doersch, Saurabh Singh, Abhinav Gupta, Josef Sivic, and Alexei~A. Efros,
\newblock ``What makes paris look like paris?,''
\newblock {\em Commun. {ACM}}, vol. 58, no. 12, pp. 103--110, 2015.

\bibitem{DBLP:journals/corr/abs-1803-07422}
Ugur Demir and G{\"{o}}zde~B. {\"{U}}nal,
\newblock ``Patch-based image inpainting with generative adversarial
  networks,''
\newblock {\em CoRR}, vol. abs/1803.07422, 2018.

\bibitem{DBLP:conf/iccv/HanWHS019}
Xintong Han, Zuxuan Wu, Weilin Huang, Matthew~R. Scott, and Larry Davis,
\newblock ``Finet: Compatible and diverse fashion image inpainting,''
\newblock in {\em {ICCV}}. 2019, pp. 4480--4490, {IEEE}.

\bibitem{DBLP:conf/icmcs/LiuGCY019}
Sen Liu, Zongyu Guo, Jiale Chen, Tao Yu, and Zhibo Chen,
\newblock ``Interleaved zooming network for image inpainting,''
\newblock in {\em {ICME} Workshops}. 2019, pp. 673--678, {IEEE}.

\bibitem{ma2019coarse}
Yuqing Ma, Xianglong Liu, Shihao Bai, Lei Wang, Dailan He, and Aishan Liu,
\newblock ``Coarse-to-fine image inpainting via region-wise convolutions and
  non-local correlation,''
\newblock in {\em {IJCAI}}. 2019, pp. 3123--3129, ijcai.org.

\bibitem{DBLP:conf/nips/WangTQSJ18}
Yi~Wang, Xin Tao, Xiaojuan Qi, Xiaoyong Shen, and Jiaya Jia,
\newblock ``Image inpainting via generative multi-column convolutional neural
  networks,''
\newblock in {\em NeurIPS}, 2018, pp. 329--338.

\bibitem{xiong2019foreground}
Wei Xiong, Jiahui Yu, Zhe Lin, Jimei Yang, Xin Lu, Connelly Barnes, and Jiebo
  Luo,
\newblock ``Foreground-aware image inpainting,''
\newblock in {\em {CVPR}}. 2019, pp. 5840--5848, Computer Vision Foundation /
  {IEEE}.

\bibitem{DBLP:conf/nips/GoodfellowPMXWOCB14}
Ian~J. Goodfellow, Jean Pouget{-}Abadie, Mehdi Mirza, Bing Xu, David
  Warde{-}Farley, Sherjil Ozair, Aaron~C. Courville, and Yoshua Bengio,
\newblock ``Generative adversarial nets,''
\newblock in {\em {NIPS}}, 2014, pp. 2672--2680.

\bibitem{DBLP:journals/tog/IizukaS017}
Satoshi Iizuka, Edgar Simo{-}Serra, and Hiroshi Ishikawa,
\newblock ``Globally and locally consistent image completion,''
\newblock {\em {ACM} Trans. Graph.}, vol. 36, no. 4, pp. 107:1--107:14, 2017.

\bibitem{DBLP:conf/cvpr/YangLLSWL17}
Chao Yang, Xin Lu, Zhe Lin, Eli Shechtman, Oliver Wang, and Hao Li,
\newblock ``High-resolution image inpainting using multi-scale neural patch
  synthesis,''
\newblock in {\em {CVPR}}. 2017, pp. 4076--4084, {IEEE} Computer Society.

\bibitem{DBLP:conf/eccv/LiuRSWTC18}
Guilin Liu, Fitsum~A. Reda, Kevin~J. Shih, Ting{-}Chun Wang, Andrew Tao, and
  Bryan Catanzaro,
\newblock ``Image inpainting for irregular holes using partial convolutions,''
\newblock in {\em {ECCV} {(11)}}. 2018, vol. 11215 of {\em Lecture Notes in
  Computer Science}, pp. 89--105, Springer.

\bibitem{DBLP:journals/corr/abs-2108-09760}
Xiefan Guo, Hongyu Yang, and Di~Huang,
\newblock ``Image inpainting via conditional texture and structure dual
  generation,''
\newblock {\em CoRR}, vol. abs/2108.09760, 2021.

\bibitem{DBLP:conf/eccv/YanLLZS18}
Zhaoyi Yan, Xiaoming Li, Mu~Li, Wangmeng Zuo, and Shiguang Shan,
\newblock ``Shift-net: Image inpainting via deep feature rearrangement,''
\newblock in {\em {ECCV} {(14)}}. 2018, vol. 11218 of {\em Lecture Notes in
  Computer Science}, pp. 3--19, Springer.

\bibitem{DBLP:conf/iccv/LiuJX019}
Hongyu Liu, Bin Jiang, Yi~Xiao, and Chao Yang,
\newblock ``Coherent semantic attention for image inpainting,''
\newblock in {\em {ICCV}}. 2019, pp. 4169--4178, {IEEE}.

\bibitem{DBLP:conf/cvpr/HuSS18}
Jie Hu, Li~Shen, and Gang Sun,
\newblock ``Squeeze-and-excitation networks,''
\newblock in {\em {CVPR}}. 2018, pp. 7132--7141, Computer Vision Foundation /
  {IEEE} Computer Society.

\bibitem{Li_2019_CVPR}
Xiang Li, Wenhai Wang, Xiaolin Hu, and Jian Yang,
\newblock ``Selective kernel networks,''
\newblock in {\em Proceedings of the IEEE/CVF Conference on Computer Vision and
  Pattern Recognition (CVPR)}, June 2019.

\bibitem{gottlob:nonmon}
Georg Gottlob,
\newblock ``Complexity results for nonmonotonic logics,''
\newblock {\em Journal of Logic and Computation}, vol. 2, no. 3, pp. 397--425,
  June 1992.

\bibitem{DBLP:journals/corr/YuK15}
Fisher Yu and Vladlen Koltun,
\newblock ``Multi-scale context aggregation by dilated convolutions,''
\newblock in {\em {ICLR} (Poster)}, 2016.

\bibitem{DBLP:conf/eccv/LiuJSHY20}
Hongyu Liu, Bin Jiang, Yibing Song, Wei Huang, and Chao Yang,
\newblock ``Rethinking image inpainting via a mutual encoder-decoder with
  feature equalizations,''
\newblock in {\em {ECCV} {(2)}}. 2020, vol. 12347 of {\em Lecture Notes in
  Computer Science}, pp. 725--741, Springer.

\bibitem{DBLP:conf/iclr/Jolicoeur-Martineau19}
Alexia Jolicoeur{-}Martineau,
\newblock ``The relativistic discriminator: a key element missing from standard
  {GAN},''
\newblock in {\em {ICLR} (Poster)}. 2019, OpenReview.net.

\bibitem{DBLP:conf/iccvw/NazeriNJQE19}
Kamyar Nazeri, Eric Ng, Tony Joseph, Faisal~Z. Qureshi, and Mehran Ebrahimi,
\newblock ``Edgeconnect: Structure guided image inpainting using edge
  prediction,''
\newblock in {\em {ICCV} Workshops}. 2019, pp. 3265--3274, {IEEE}.

\end{thebibliography}

\end{document}